# When Gravity Fails: Local Search Topology

**Jeremy Frank**  FRANK@TIZIANO.ARC.NASA.GOV
*Caelum Research Corp.*
*NASA Ames Research Center*
*Mail Stop N269-1*
*Moffett Field, CA 94035-1000*

**Peter Cheeseman**  CHEESEM@PTOLEMY.ARC.NASA.GOV
*RIACS*
*NASA Ames Research Center*
*Mail Stop N269-1*
*Moffett Field, CA 94035-1000*

**John Stutz**  STUTZ@PTOLEMY.ARC.NASA.GOV
*NASA Ames Research Center*
*Mail Stop N269-1*
*Moffett Field, CA 94035-1000*

## Abstract

Local search algorithms for combinatorial search problems frequently encounter a sequence of states in which it is impossible to improve the value of the objective function; moves through these regions, called *plateau moves*, dominate the time spent in local search. We analyze and characterize *plateaus* for three different classes of randomly generated Boolean Satisfiability problems. We identify several interesting features of plateaus that impact the performance of local search algorithms. We show that local minima tend to be small but occasionally may be very large. We also show that local minima can be escaped without unsatisfying a large number of clauses, but that systematically searching for an escape route may be computationally expensive if the local minimum is large. We show that plateaus with exits, called benches, tend to be much larger than minima, and that some benches have very few exit states which local search can use to escape. We show that the solutions (i.e., global minima) of randomly generated problem instances form clusters, which behave similarly to local minima. We revisit several enhancements of local search algorithms and explain their performance in light of our results. Finally we discuss strategies for creating the next generation of local search algorithms.

## 1. Introduction

Local search algorithms have been heavily studied as an alternative to complete search for $\mathcal{NP}$-Hard problems. A typical local search algorithm, such as gradient descent or greedy search, employs an objective function to rank states, and picks a "neighboring" state maximizing the improvement to the objective function. A compelling (if inexact) analogy is that of dropping a marble on a smooth surface and observing it roll downhill into a local valley. The typical greedy objective function acts like gravity, pulling the current state downhill. This procedure can result in the algorithm becoming trapped in a local minimum. Local search algorithms tend to find solutions to satisfiable decision problems more quickly than complete search algorithms. However, these algorithms may terminate

```
procedure GSAT(Σ,MaxFlips, MaxTries)
# Σ is the problem instance to be solved
# A is the current variable assignment
   for i=1 to MaxTries
     A = N-bit string selected uniformly at random
     for j = 1 to MaxFlips
       if solved_problem(A, Σ)
         return A
       else
         PossFlips = neighbors of A minimizing the number of unsatisfied clauses
         A = one element of Possflips selected uniformly at random
       end else
     end for
   end for
   return FAIL
end
```

Figure 1: GSAT Algorithm Sketch

either without finding a solution when one exists or guaranteeing that a problem instance does not have a solution.

GSAT is a local search procedure for the Boolean Satisfiability problems (Selman, Levesque, & Mitchell, 1992) that has proven to be effective at quickly finding solutions to satisfiable problem instances. A sketch of GSAT appears in Figure 1. In the figure, $\Sigma$ refers to the problem instance GSAT is to solve. GSAT's search space is the space of all complete assignments of values to variables. The GSAT algorithm is typically given a fixed number of tries (denoted in the figure as *MaxTries*) and a fixed number of moves per try (denoted *MaxFlips*) to solve a problem instance. During each move, GSAT examines all states reachable by changing the value of a single variable, and selects moves that minimize the number of unsatisfied clauses. GSAT typically encounters a sequence of states where the best move available at each state leaves the number of unsatisfied clauses unchanged. These moves are referred to as *plateau moves* or *sideways moves*, studied in (Gent & Walsh, 1993a) and (Hampson & Kibler, 1995). Plateau moves dominate the time GSAT spends doing search (Gent & Walsh, 1993a). It is believed that all combinatorial search problems with discrete objective functions have *plateaus* that cause plateau moves during local search, but it is unlikely that search problems with real-valued objective functions have plateaus. When GSAT encounters a plateau, it randomly searches until it either runs out of flips or finds a neighboring state with fewer unsatisfied clauses, thereby exiting the plateau. Returning to the marble analogy, there is no gravity on the plateaus, and hence the marble simply rolls at random until it finds an exit or runs out of momentum. Numerous variants of GSAT have been developed to avoid random plateau search and improve GSAT performance (Gent & Walsh, 1993b; Selman & Kautz, 1993; Gent & Walsh, 1995).

The nature of plateau behavior of local search algorithms is not well understood. Some researchers suggest that algorithms like GSAT become trapped in *local minima*, i.e., parts of the search space from which there is no escape to a better part of the search space. If this is true, local minima detection and avoidance is the most important problem in local search algorithm development. Other researchers have suggested that local search could become trapped in "flat" regions of the search space that have exits to better states, which we call *benches*. This may happen because benches are large, or because they contain few exits and random plateau search has a small probability of finding an exit. Rather than designing algorithms and testing them on problem classes, we undertook an empirical examination of the nature of plateaus for a variety of 3-SAT problems.

This paper presents several surprising discoveries concerning the topological structures leading to plateau behavior and their impact on local search. We define *plateaus* as a feature of the search space and break plateaus into two classes: *local minima* and *benches*. Plateaus are defined as any maximally connected region of the local search space over which the objective function is constant. Local minima are plateaus surrounded by regions of the search space where the objective function takes on values exceeding that of the plateau, with the result that purely greedy local search cannot escape once finding a state on the local minimum. Benches are defined as plateaus with exits to regions of the search space with lower values of the objective function. Our results show that local minima are more common than benches when the number of unsatisfied clauses is close to 1, but local minima also occur frequently at higher numbers of unsatisfied clauses. Most local minima tend to be small, but their size exhibits high variability; often the largest local minima exceeds 10,000 states in a problem instance containing 100 variables. Also surprising was the behavior of solutions: solutions are grouped together into global minima of highly variable size. Our results also show that benches tended to be much larger than local minima. Most benches have a large number of exits, but a small fraction have very few exits, with the result that local search can spend a large amount of time trying to escape them. Plateau characteristics are dependent on many features of a problem instance; we found differences in plateau characteristics based on the ratio of clauses to variables, solvability of the problem instances, and problem classes. The results on plateau characteristics allowed us to reinterpret the success of many modifications to local search, including history lists (Gent & Walsh, 1993b), random walk (Kautz & Selman, 1996) and tabu search (Glover, 1989).

The paper is organized as follows: In Section 2 we present some definitions used throughout the rest of the paper. Next in Sections 3 and 4 we present an empirical analysis of properties of plateaus for several problem spaces. In Section 5 we present an analysis of previous results in light of our findings. We then suggest how to apply our work to the creation of new local search algorithms in Section 6, and finally in Section 7 we conclude and discuss ideas for future work.

## 2. Definitions

In this section we will define some terms used throughout the paper. We restrict our discussion to the Boolean Satisfiability problems in conjunctive normal form with three distinct literals per clause, abbreviated 3-SAT, but many of the concepts presented here translate to other discrete combinatorial search problems. We first present informal definitions, and provide more formal definitions at the end of the section.

A way of visualizing the local search space for 3-SAT is by mapping each full variable assignment to a node of an $N$ dimensional hypercube, where $N$ is the number of variables in the problem instance. If two assignments differ by one variable assignment they are adjacent nodes in the hypercube. Each problem instance $\mathcal{I}$ defines a function on nodes of the hypercube, mapping the node to the number of unsatisfied clauses of the instance under the assignment of values corresponding to the node. We refer to the number of unsatisfied clauses under an assignment as the *level* of the assignment. A *plateau* is a maximal connected region of the assignment space where all states have the same level, and the level of the plateau is the level of the states in the plateau. Even a single state can be a plateau, if all of its neighbors are of a different level than the state itself. We define the *border* of a plateau to be the set of nodes in the hypercube that are neighbors of some state on the plateau but have a different level than the plateau. A plateau is a *minimum* if all states on the border have a higher level than the plateau. If a plateau is not a minimum, then there is some state on the border with a lower level than the states on the plateau; states on the plateau that are adjacent to these lower level states are called *exits*. Plateaus with exits are called *benches*. Some benches consist entirely of exits; a local search algorithm may then explore only one state of such a bench before moving off of it. We call these benches *contours*. For 3-SAT, a plateau is a global minimum if it is of level 0, but unsatisfiable problem instances can have global minima of levels higher than 0.

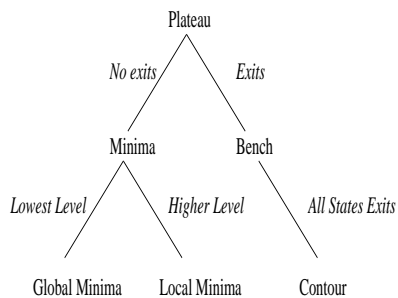

Figure 2: A Taxonomy of Plateaus

In summary: A plateau is a part of the space that is "flat" from the perspective of the objective function. All the states neighboring the plateau are of a different level from the plateau. If all the neighboring states are at a higher level, the plateau is a local or global minimum, otherwise it is a bench. If every state on the plateau is a neighbor of a state of lower level, the bench is a contour. Figure 2 shows a taxonomy of different types of plateaus. A further illustration of these definitions using a simple problem instance is presented in Appendix A.

We realize that there are different ways of defining topological structures of local search spaces. Our definition of plateaus includes structures which do not lead to plateau behavior; local search will not exhibit plateau behavior as it passes a contour, for example. These definitions show that the observed plateau behavior of local search can be caused by a variety of structures in the local search topology. In retrospect, it is clear that a definition of benches that specifically excludes contours would better serve to characterize the plateau behavior of greedy algorithms. We caution the reader that our results for benches are contingent upon our current definition of benches, and that there is at least one reasonable alternate definition that is expected to give somewhat different results. This will be discussed further in Section 7.

We end this section by providing more formal definitions of these ideas. Throughout the following definitions, let $H$ be an $N$ dimensional hypercube representing the possible assignments of a 3-SAT problem instance $\mathcal{I}$. Two vertices of the hypercube $h_1, h_2$ are neighbors, i.e., have an edge between them, if they correspond to assignments differing in exactly 1 variable.

**Definition 2.1 (Level)** *Let $\mathcal{I} : H \to \mathcal{Z}^+$ be a function mapping assignments to integers such that $\mathcal{I}(h) = z$ if and only if the assignment corresponding to h results in z unsatisfied clauses in problem instance $\mathcal{I}$. Then z is defined as the **level** of the assignment.*

**Definition 2.2 (Plateau)** *Let P be a connected subgraph of H and let $z \in \mathcal{Z}^+$ be a constant. Then P is a **plateau** if P is a maximal connected subgraph of H such that $\mathcal{I}(p) = z$ for all $p \in P$. Further, z is defined to be the level of the plateau.*

**Definition 2.3 (Border)** *Let P be a plateau in a hypercube H. Let N(p) be the set of neighboring vertices of vertex p in the hypercube. Let V(P) be a function that returns the vertex set of a graph P. Define $B(P) = (\cup_{p \in P} N(p)) - V(P)$, i.e., the set B(P) contains b if b is a neighbor of a vertex $p \in P$ and b is not in P itself. Then B(P) is the **border** of the plateau.*

**Definition 2.4 (Minimum, Local Minimum, Global Minimum)** *Let P be a plateau in a hypercube H. Then P is a **minimum** if all vertices in B(P) have higher level than the level of P. Also, P is a **local minimum** if P is a minimum and there is another minimum Q such that the level of Q is smaller than the level of P. If P is a minimum that is not a local minimum then P is a **global minimum**.*

**Definition 2.5 (Bench, Exit, Contour)** *Let P be a plateau of a hypercube H. Then P is a **bench** if P is not a minimum, and hence there exists $b \in B(P)$ such that the level of b is smaller than the level of P. Also, p is an **exit** from the bench if $p \in P$ and p is a neighbor of $b \in B(P)$ such that the level of b is smaller than the level of P. Finally, P is a **contour** if every state of P is an exit from P.*

## 3. Probabilistically Painting Plateaus

Armed with the definitions from the previous section we examined the landscape of plateaus for randomly generated 3-SAT problem instances. We generated problem instances for which the ratio of the number of clauses $C$ to the number of variables $N$ ranged from 3.8 to 4.6 according to the Uniform3-SAT problem generation model (Selman et al., 1992; Crawford & Auton, 1993); the algorithm for generating these instances is presented in Appendix B. Problems in this region straddle the "phase transition" in satisfiability, for which the satisfiability of randomly generated problems exhibits a rapid transition with respect to the ratio of clauses to variables, and for which complete search and GSAT require the longest time on the average to find solutions (Cheeseman, Kanefsky, & Taylor, 1991; Crawford & Auton, 1993; Clark, Frank, Gent, MacIntyre, Tomov, & Walsh, 1996). Problem instances with $\frac{C}{N} < 4.3$ are referred to as "under-constrained" since they lie below the observed transition in satisfiability, while problem instances with $\frac{C}{N} > 4.3$ are called "over-constrained." We guaranteed that each problem instance used in this set of experiments was satisfiable by finding a solution using a complete search algorithm.

Local search seems to have the most difficulty when the level of the assignment becomes close to 0; consequently, we decided to analyze plateaus at these levels. It is quite difficult to randomly sample plateaus of a fixed level for a problem instance; the probability of randomly generating an assignment with one unsatisfied clause, for instance, is very small for problem instances with 100 variables. We used GSAT to find the plateaus analyzed in this paper. This biases our investigation of plateaus to those found by one local search method, but hopefully provides a first picture of the plateau structure of local search spaces. Due to the clumsiness of language, we do not remind the reader throughout the text that our findings are dependent on our plateau sampling methodology. Further, because GSAT employs random starting points, the bias of our results depends only on the gradient following procedure. To sample plateaus we first used GSAT to find a state of a pre-determined level. That is, we generated an initial state and ran a single try of GSAT until it encountered a state with the specified number of unsatisfied clauses. We then used Breadth-First Search to find all of the states on the plateau found by GSAT. Naturally Breadth-First Search records each state found so that redundant states are not double-counted. We then recorded the size of the plateau (i.e., the number of states on the plateau), and the number of exits the plateau contained.

### 3.1 Characterizing Plateaus

We first analyzed the relative proportions of benches and minima of satisfiable problem instances for plateaus whose level was close to 0. We generated problem instances of 100 variables and 380-460 clauses in increments of 10 clauses. For each problem size we generated 1000 problem instances and guaranteed each instance had a solution using a complete search algorithm. Using the procedure described above, for each problem instance generated we found one plateau of each level from 0 to 5 and measured the proportion of these plateaus that are local minima and benches. This analysis does not provide any idea of the number of benches or local minima in these problem instances. Note that all plateaus of level 0 are global minima of satisfiable problem instances.

Figure 3 shows the proportion of plateaus that are local minima graphed against the number of clauses in the problem instances. As described above, we used GSAT to find plateaus and Breadth-First Search to determine whether the plateaus were local minima or benches. Here we see that the proportion of plateaus that are minima grows with the number of clauses in the problem instance for plateaus of levels 2-5; hence there are more local minima of identical levels in over-constrained problems than in under-constrained problems. The rate of growth diminishes as the plateau level decreases, until it is roughly flat for plateaus of level 1. About 85% of plateaus of level 1 are minima for 100 variable problem instances over all numbers of clauses investigated.

Figures 4 shows the same data, except in this case we have graphed against the level of the plateau. As the level grows the proportion of local minima declines for problem instances of all numbers of clauses. However, plateaus at level 5 may still be local minima

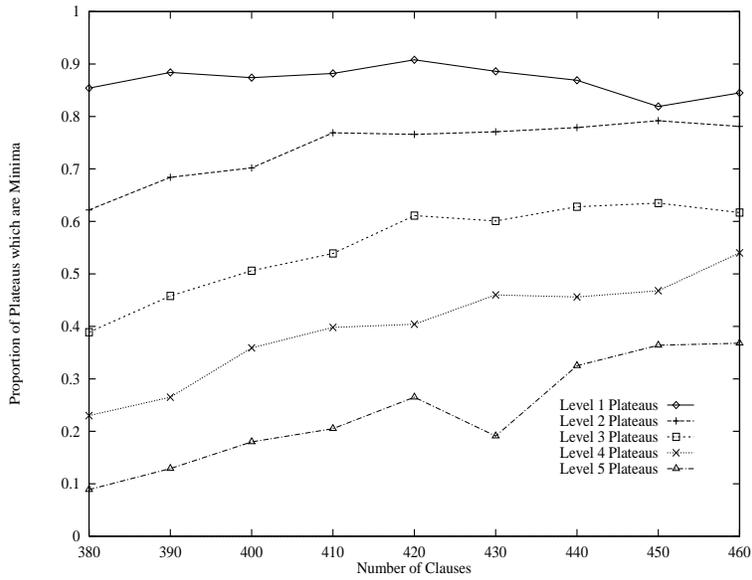

Figure 3: Proportion of Plateaus that are Local Minima vs. Number of Clauses for Randomly Generated 100 Variable Problem Instances

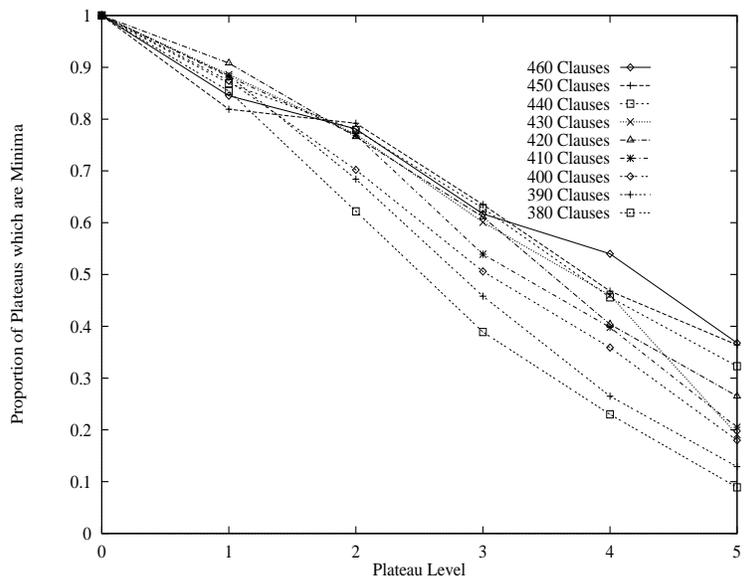

Figure 4: Proportion of Plateaus that are Local Minima vs. Plateau Level for 100 Randomly Generated Variable Problem Instances With Varying Numbers of Clauses

even for problems with 100 variables and 380 clauses. Hence local minima are a fact of life even for under-constrained problems, and become more likely for over-constrained problems. Finally, we note that between plateaus of level 1 and 2 there is a reordering of the proportion of local minima. For example, problems with 450 clauses have the lowest proportion of local minima of level 1, but the highest proportion of local minima of level 2. We do not have an explanation for this result.

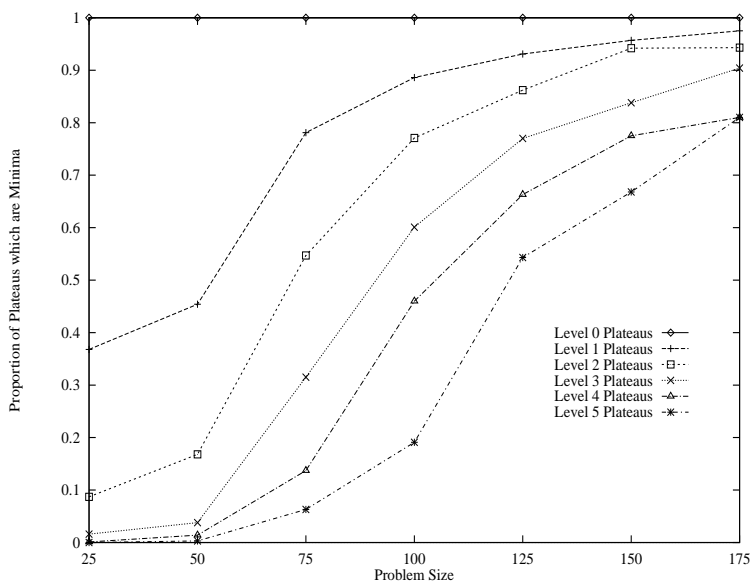

Figure 5: Proportion of Plateaus that are Local Minima vs. Plateau Level for Variable Sized Randomly Generated Problem Instances with C/N=4.3

We also analyzed how the relative proportions of benches to minima changed as problems grew larger. We found one plateau each of levels 1-5 for each of 1000 problem instances with 25 to 175 variables in increments of 25, with $\frac{C}{N} = 4.3$. Figure 5 shows the proportion of plateaus of levels 1-5 that are local minima for $\frac{C}{N} = 4.3$ graphed against the problem size. We see that, for each level, the proportion of minima grows with the problem size, which bodes ill for the performance of local search on larger problem instances. We see that the proportion of local minima of higher level decreases less rapidly for these larger problems. We conjecture that, for larger problems, the proportion of local minima decreases significantly for plateaus of levels higher than 5, but we cannot predict the exact behavior from the data at hand.

The cost of detecting local minima is proportional to the size of the local minima, so understanding the size distribution of local minima is important. The cost of escaping benches is dependent on the size of the bench and the proportion of the states on a bench that are exits, and so understanding these properties is also important. In the next two sections we analyze these characteristics of benches and minima. To do so, we generated statistics from the plateaus we found in the experiment used to generate Figure 4. For instance, 60% of the plateaus of level 2 for problem instances of 380 clauses are local minima, so we had 600 local minima and 400 benches to analyze the characteristics of plateaus. In all cases we had over 100 data points available for analysis.

### 3.2 Minima Characteristics

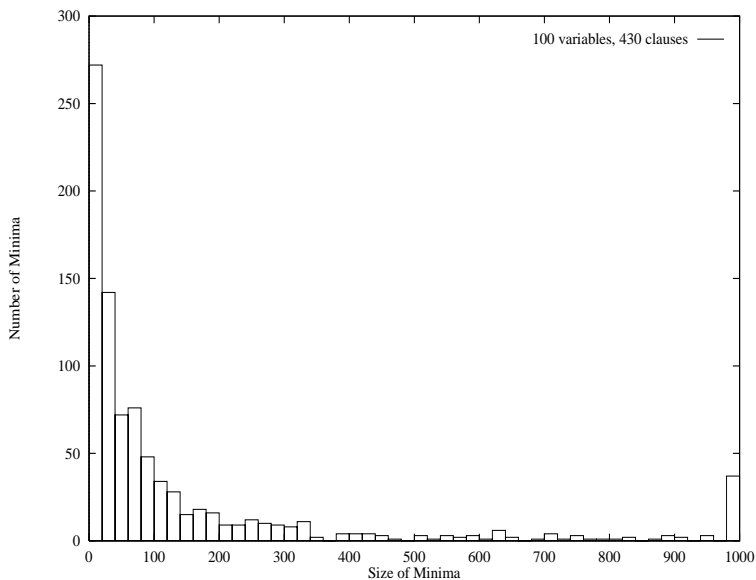

Figure 6: Histogram of Sizes of Level 1 Minima for Randomly Generated Problem Instances of 100 Variables, 430 Clauses. Note that some minima exceeded 1000 states.

In this section we analyze the size distribution of local minima. Figure 6 shows the distribution of the size of the level 1 minima for problem instances with 100 variables and 430 clauses. The median minima size is 48, yet the tail shows that some minima are larger than 1000 states. In fact, 35 of the 900 minima are larger than 1000 states and some had as many as 10,000 states. We examined the distribution of minima sizes for levels other than 1 and found similar results; the main differences are in the lengths of the tails of the histograms. A consequence of this discovery is that escaping local minima by explicit local minima detection is normally very easy, but occasionally can be very expensive. Figure 7 shows the distribution of sizes of minima that are smaller than 100 states. We see here that there are fewer minima of size 0-5 than of size 5-10; a detailed analysis reveals that there are three minima of size 1, and fifteen minima of size 2.

Due to the long tails of the distribution of minima size, the median provides a more stable summary statistic than the mean. We therefore examined the median size of local minima of levels 0-5 for different numbers of clauses to determine how the size of local minima varies. Figure 8 shows the median size of local minima plotted against the number of clauses in the problem instances. The most striking feature of these results is that most minima tend to be quite small. This suggests it is possible to devise local search algorithms to detect local minima using exhaustive search and then propel themselves into a more fruitful part of the search space. While the distribution shown in Figure 6 indicates that some local minima can be very large, explicit detection of minima below a fixed size (the median, for instance) may be a successful addition to local search. The second feature of note is that the median size of level 0 minima (i.e., solutions) follows the same pattern as local minima, but that level 0 minima tend to be smaller than local minima. The last feature of note is that the median size of local minima decreases for minima of level 0-3 as the number of clauses in the problem instances increases. The median size of local minima of level 4 and 5 increases for problems with 450-460 clauses. One possible explanation for this is that the sampled problem instances were guaranteed to have solutions, which means that the added clauses must contribute to larger minima and benches at higher levels. If this is true it is not clear why the minima of levels 0-3 do not increase in size. Another possible explanation is the small amount of data for plateaus of levels 4 and 5 relative to the amount available for the other plateau sizes as indicated in Figure 4.

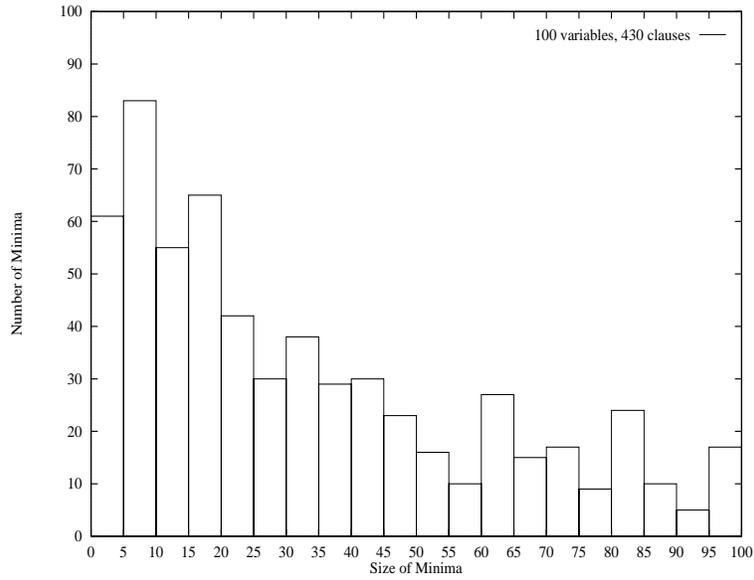

Figure 7: Histogram of Sizes of Level 1 Minima Smaller than 100 States for Randomly Generated Problem Instances of 100 Variables, 430 Clauses

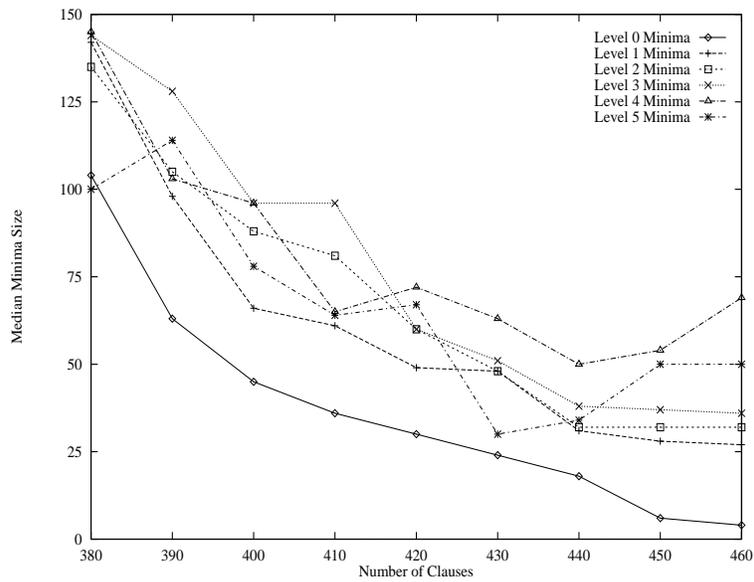

Figure 8: Median Size of Minima vs. Problem Size for 100 Variable Randomly Generated Problem Instances with Varying Numbers of Clauses

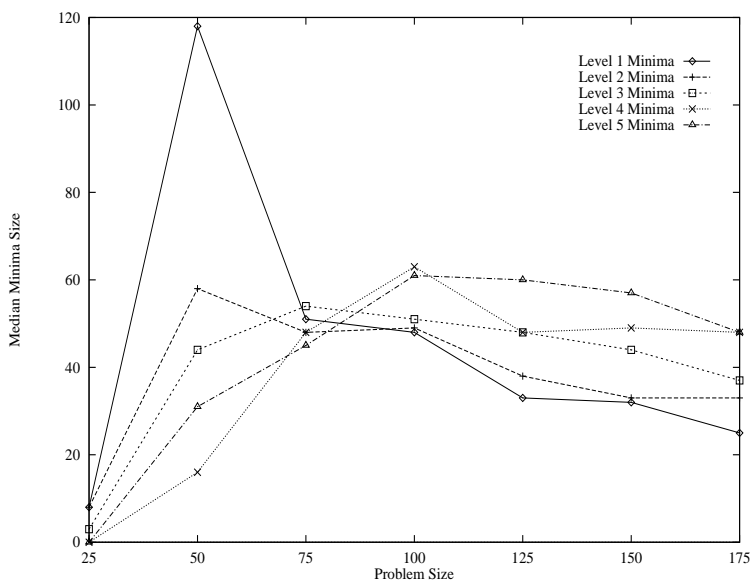

Figure 9: Median Size of Minima vs. Plateau Level for Variable Sized Randomly Generated Problem Instances with C/N=4.3

We also examined the median size of local minima of levels 1-5 as the number of variables in the problem instances increases. Figure 9 shows the median minima size for the various levels of minima graphed against problem size for problem sizes from 25 to 175 variables for $\frac{C}{N} = 4.3$. We observe that, as problem sizes grow large (beyond 100 variables), the median size of minima of lower levels appears smaller than that of minima of higher levels. We also observe that for fixed minima level, there appears to be a problem instance size that maximizes the median minima size. We do not have an explanation for the large number of minima of level 1 for problem instances of 50 variables.

Recall that a global minima for a satisfiable problem instance is a plateau where all states are solutions. How many global minima are there in satisfiable problem instances? Is there only one, or are the solutions broken into multiple global minima? If there is more than a single global minima, can the size of the global minima tell us anything about how likely it is that local search will encounter a particular solution? To answer these questions we used GSAT to find 1000 global minima for a single problem instance with 100 variables and 430 clauses, and determined which minima were distinct. We found that 436 of the 1000 minima were unique, and that the global minima for this instance ranged in size from 1 to 2880 states. Furthermore, we found that the vast majority of the minima are small, with the median size being around 48. We repeated the experiment for 20 more problem instances and found that solutions for these problem instances are typically divided into separate global minima and that the global minima vary widely in size. Due to space considerations we do not present this data in the paper.

Assuming we could detect local minima, how difficult is it for local search to escape minima in order to explore a new part of the search space? If local search is in a local minimum it must temporarily move to states of higher level in order to find a more promising part of the search space. Two sources of computational complexity contribute to the cost of escaping minima: the cost of detecting the minimum, and the cost of finding a path to a better part of the search space. The size of the minimum is a measure of the detection cost; we chose the minimum increase in level as a measure of the difficulty of escaping local minima. To understand this idea, consider all sequences of neighboring states out of a minimum such that the level increases, then decreases. We are interested in the minimum increase required before the level decreases again. To determine this we generated 1000 problem instances of 50 variables and 215 clauses each, generated an initial state, and then ran GSAT for 1000 flips. This was sufficient to reach a local or global minimum. [1] In order to find the minimum level required to escape, we used Breadth-First Search. We begin with the states of the local minimum. We then explore a state on the border, queuing all those states not explored before in increasing order of level. This ensures states of lower level are explored first. Once we encounter a state of lower level than its neighbor, we have found a path out of the local minimum; the level of the state with a neighbor of lower level is the minimum level required to escape the local minimum. Our results indicate that local minima can usually be escaped by increasing the level by only 1, that is, only unsatisfying one additional clause. However, it is not obvious *which* border state to use to escape; Breadth-First Search may expand tens of thousands of states before finding an escape route, and so this may not always be an effective escape strategy.

In summary, the data presented in Figures 6 to 9 shows that local minima tend to be very small much of the time, and therefore may be easily detectable and escaped. Local minima can typically be escaped by unsatisfying only a single additional clause, but it is still not clear how to escape local minima effectively. Further, the size distribution of global minima behaves much like the size distribution of local minima. Instances tend to have many global minima of highly variable size, and there is evidence that local search is more likely to encounter small sized local minima and global minima than large ones.

### 3.3 Bench Characteristics

---

1. The local minima ranged in level from 1 to 6; we did not measure the level required to escape from benches or global minima.

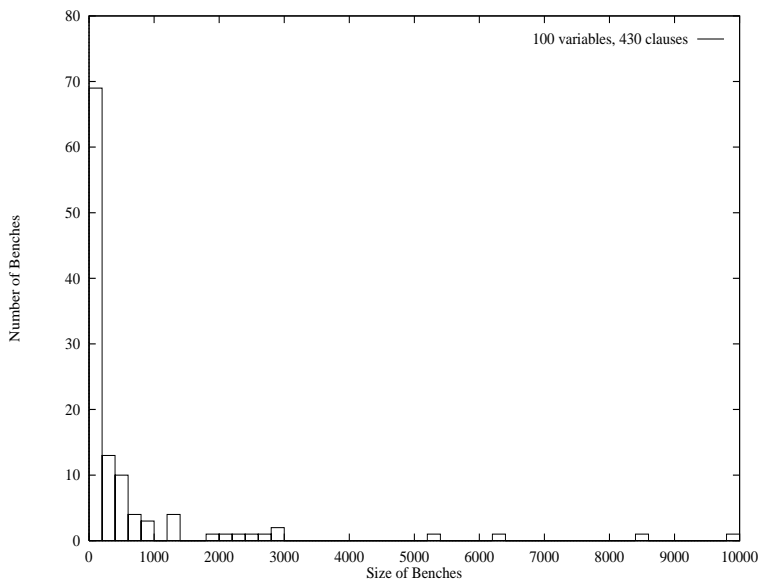

Figure 10: Histogram of Bench Sizes of Level 1 for Randomly Generated Problem Instances of 100 Variables, 430 Clauses

Recall that a bench is a plateau that has exits to other states of lower level. Two important characteristics of benches that impact the performance of local search are the size of the benches and the number of exits. We first analyzed the distribution of the size of benches; Figure 10 shows the distribution of bench sizes of level 1 for problem instances of 100 variables and 430 clauses. Again we found long tails, implying that while most benches are small, some can be very large. The distributions tend to be much flatter than those for minima.

We next analyzed how median bench size varies with the number of clauses in the problem. The appearance of long tails of the distribution of bench sizes again indicates that the median is a more stable measure than the mean. Figure 11 shows how the median size of benches varies with the number of clauses in problems for different levels of benches. The most interesting feature is the very large median size of the benches when compared to the size of local minima. Benches are typically 10-30 times as large as local minima, depending on the level and number of clauses in the problem instance. Problem instances with more clauses tend to have smaller benches, while for some of the under-constrained instances the median bench size begins growing rapidly with bench level even for the small range of plateau levels analyzed here.

We also examined how bench size depends on problem size. For small problems, i.e., 25-50 variables, benches tended to be much larger than for problems with more variables. Our explanation for this is that there are so few clauses that they do not adequately distinguish between assignments when we examine states at low enough levels. Problems with 100

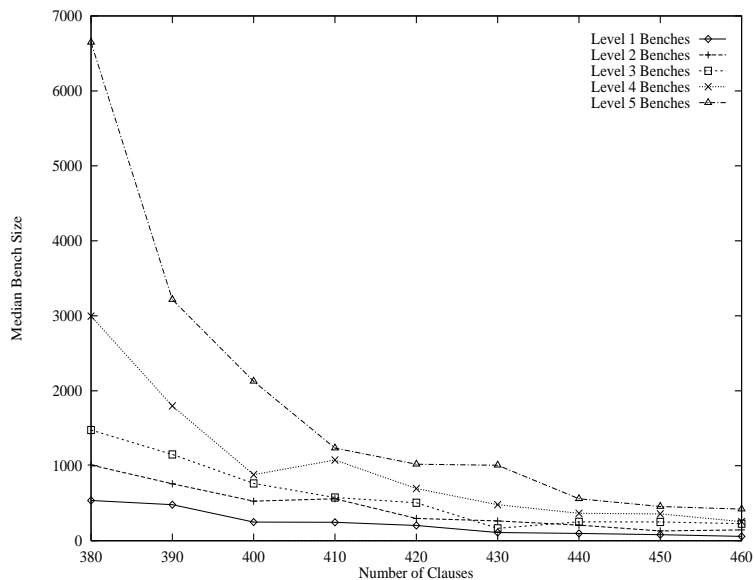

Figure 11: Median Size of Benches vs. Problem Size for 100 Variable Randomly Generated Problem Instances

variables have few benches with more than 10,000 states; we exclude these benches from our analysis. [2]

Large benches may be difficult to escape if the number of exits is small, or if exits are clustered together. Some benches have very few exits, while others have many exits. We used the ratio of exits from benches to the bench size as a measure of the difficulty of escaping from benches. We did not investigate whether or not exits from benches are close together, which may also have an impact on the difficulty of escaping benches.

Figure 12 shows the distribution of proportion of exits to bench size for benches of level 1 for problems with 100 variables and 430 clauses. The distribution of these proportions indicates that plateaus have highly variable numbers of exits. We note that some benches are in fact contours; in Figure 12 contours show up as benches whose ratio of exits to bench size is 1. We observed that all six of the benches with ratio of exits to bench size at least 0.95 in Figure 12 are in fact contours. Figure 13 shows the distribution of the proportion of exits for benches of level 5; there are proportionally more contours (65 of the 78 benches in the rightmost column of the histogram are contours), and the mean ratio of exits to bench size has increased. The difference between these two plots indicates that benches of lower levels tend to have fewer exits than benches of higher level, even if we exclude contours from the measurements.

To further understand how to escape benches of different levels, we next present plots of the mean proportion of the number of exits to bench size graphed against problem size in Figure 14. When taken in consideration with the histograms in Figure 12, we hope this will create an accurate picture of how benches tend to look.

---

2. Breadth-First Search stores an enormous number of states, and as such we terminated the program if the bench size exceeded 10,000 states. Since we had so few large benches and used the median statistic, this caused few difficulties in the analysis.

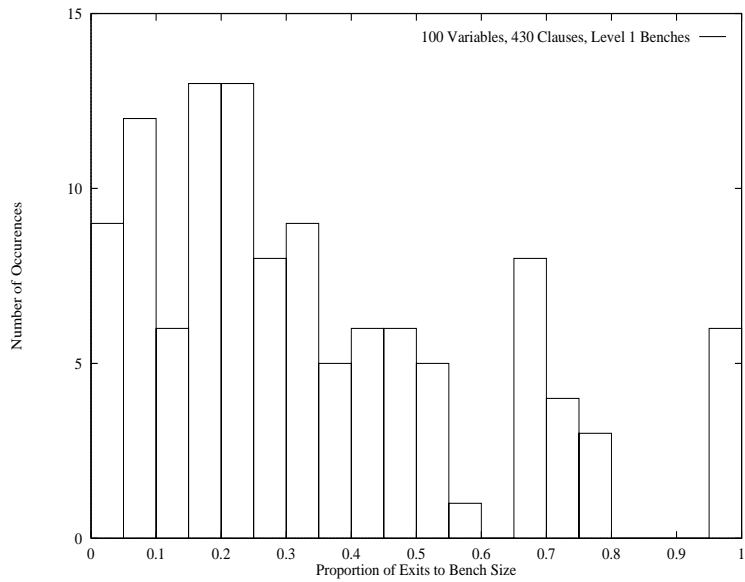

Figure 12: Exits from Level 1 Benches for Randomly Generated Problem Instances of 100 Variable, 430 Clauses

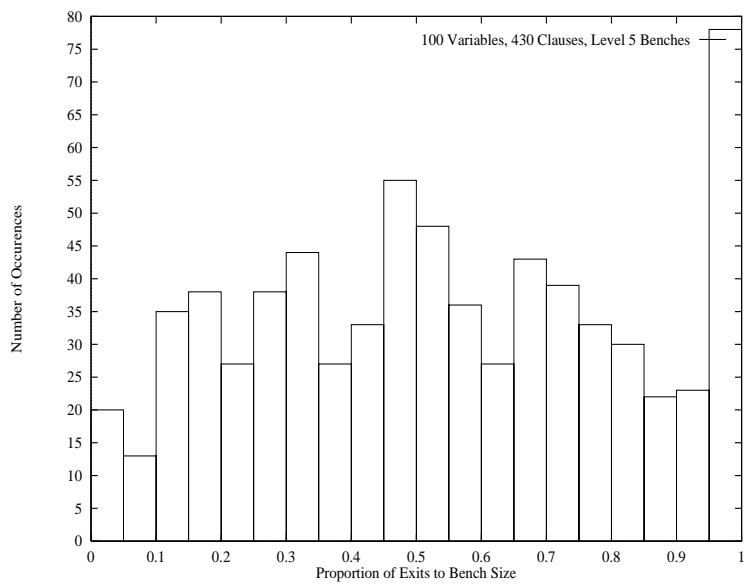

Figure 13: Exits from Level 5 Benches for Randomly Generated Problem Instances of 100 Variables, 430 Clauses

We see in Figure 14 that the proportion of exits from benches increases with the level of the benches. For problems with 430 to 460 clauses the mean number of exits of benches of level 1 increases sharply, indicating that for over-constrained solvable problems benches of level 1 are less of an obstacle to finding solutions. We should point out that our inclusion of contours in the definition of benches may artificially inflate these proportions, in some cases considerably.

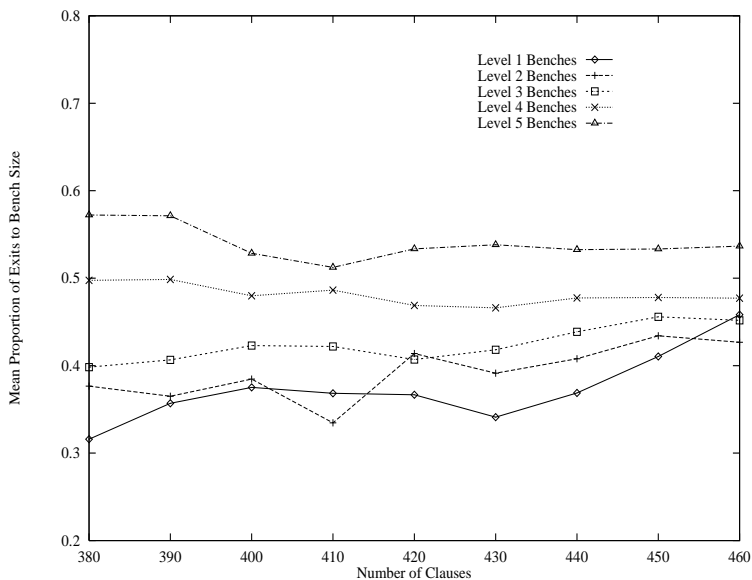

Figure 14: Mean Proportion of Exits From Benches vs Problem Size for Randomly Generated Problem Instances of 100 Variables

We analyzed benches to determine whether or not there was some relationship between the number of exits from a bench and its size, but found no such relationship for all clause sizes and benches of all levels we investigated. This lack of relationship is unfortunate, since it tells us little about how to escape large benches.

In summary, the data presented in figures 10 to 14 shows that benches are occasionally very large, but there are often many exits from benches. As a result, only occasionally will local search become trapped on a very large bench from which there is little chance to escape. We also found that benches of higher level have more exits than benches of lower level. We showed that contours are common for benches of level 5 but may also occur at level 1. Finally, we found no obvious relationship between bench size and the number of exits. We conclude that local minima are more often a problem for local search than benches since most benches seem to be easy to escape.

## 4. Plateau Characteristics Across Problem Classes

In the previous section we analyzed plateaus for satisfiable 3-SAT problem instances from one problem instance class. There is little reason to suspect that plateaus behave similarly across problem instance classes. There may also be differences between satisfiable and unsatisfiable instances of the same class. In recent years numerous algorithm designers have begun testing algorithms on random problem classes that have pre-specified desirable properties. Among these are problem instances that have guaranteed solutions (Tsuji & Gelder, 1993), and problems that have some structure that is hidden from the algorithm. A class of "cluster" problems was presented by Kask and Dechter (1995); these problems consist of a number of satisfiability problems over independent sets of variables with some number of clauses connecting the clusters. We repeated our experiments on these problem distributions to determine if plateaus in these instances exhibit different properties than the Uniform3-SAT class, and how this might alter the effectiveness of local search. We also repeated the experiments on unsatisfiable instances of the Uniform3-SAT distribution. Due to space considerations we do not repeat in full the analysis performed above, but discuss some differences between the characteristics of the classes we investigated.

### 4.1 Unsatisfiable Problems

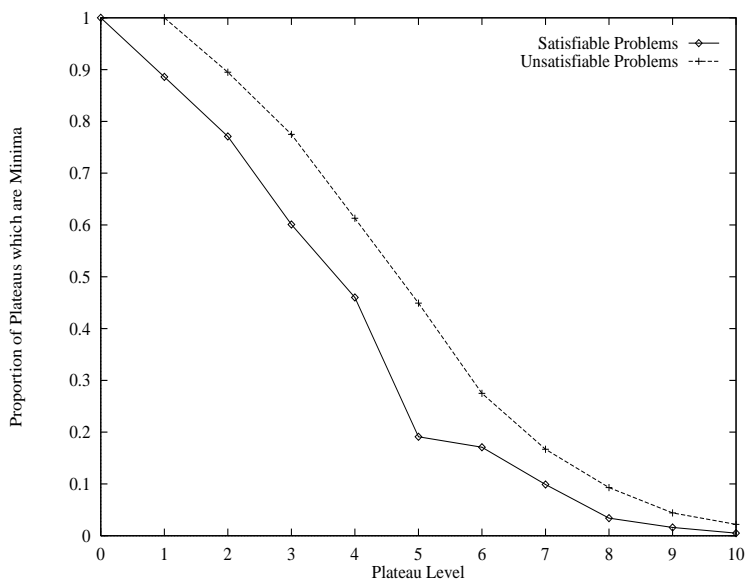

Figure 15: Proportion of Plateaus that are Local Minima for Randomly Generated Satisfiable and Unsatisfiable Problems of 100 Variables, 430 Clauses

A major drawback to local search is its inability to distinguish satisfiable problem instances from unsatisfiable problems. We analyzed the plateau structure of unsatisfiable problem instances from the Uniform3-SAT instance distribution to determine whether there are differences which would allow local search to determine that a problem instance is unsatisfiable. We repeated the previous empirical studies and collected data on the proportion and size of plateaus but limited our investigation to problems with 100 variables and 430 clauses. We first present data on the proportion of plateaus that are local minima for unsatisfiable problem instances of 100 variables and 430 clauses. We generated 1000 unsatisfiable instances using the same problem generation technique used in the previous set of experiments and guaranteed the problem instances were unsatisfiable using a complete search algorithm. We used GSAT to find states of level 1-10, then generated the corresponding plateaus. We contrast this data with the same data for satisfiable problem instances of 100 variables and 430 clauses in Figure 15.

Figure 15 shows that the proportion of plateaus that are local minima is similar for satisfiable and unsatisfiable problems as the plateau level decreases, except that for plateaus of levels 0 to 5 the proportion is shifted to the right by ome level. A possible interpretation of this result comes from noting that frequently adding a single randomly generated clause can turn a random satisfiable instance into an unsatisfiable instance. Hence local search may become trapped at a higher level in local minima for unsatisfiable problems than for satisfiable problems.

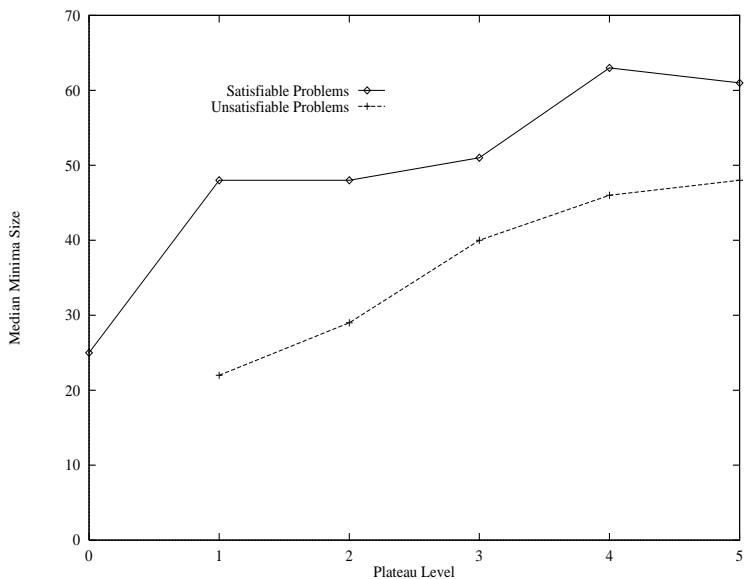

Figure 16: Median Size of Local Minima for Randomly Generated Satisfiable and Unsatisfiable Problems of 100 Variables, 430 Clauses

Next we analyzed the median size of local minima of unsatisfiable problem instances for minima of levels 1 to 5. Again we analyzed the data from the plateaus found to generate Figure 15. Since the number of minima is very small for minima of levels 6-10, we could not gather sufficient data for analysis in a reasonable amount of time. However, as Figure 16 shows, local minima for unsatisfiable problems tend to be much smaller than local minima for satisfiable problems. Figure 17 shows the median size of benches for unsatisfiable problem instances. We found that the median bench size for unsatisfiable instances tended to be smaller than benches for satisfiable problems. We conjecture that local search may converge to local minima faster for unsatisfiable problems and that the minima tend to be at a higher level; since minima and benches are smaller and there are more minima at higher levels, local search should be able to descend faster on the average and become stuck earlier. Unfortunately, the differences are slight enough that there seems little hope of using these results to improve the ability of local search to identify unsatisfiable problem instances.

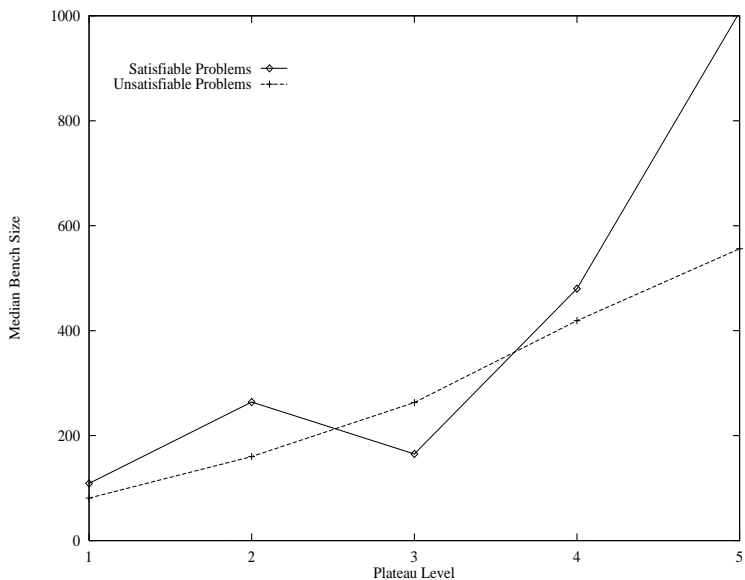

Figure 17: Median Size of Benches for Randomly Generated Satisfiable and Unsatisfiable Problems of 100 Variables, 430 Clauses

### 4.2 Instances With Guaranteed Solutions

We next present data on problems with guaranteed solutions as described in (Tsuji & Gelder, 1993). This generator is the HardSolvable3-SAT generator presented in Appendix B. Briefly, this generator selects an assignment to be a guaranteed solution, then during generation rejects both clauses that are not satisfied by the assignment and a set of additional clauses that enforce an even distribution of signs for each variable. As before, we generated 1000 problem instances, each with 100 variables and 380-460 clauses. For each problem instance we used GSAT to find a state with 1-5 unsatisfied clauses, then determined whether the corresponding plateau was a bench or a local minimum. Figure 18 shows the proportion of plateaus that are minima graphed against the number of clauses in problem instances of 100 variables. The proportion of local minima for these problems is similar but not identical to the proportions of local minima for the Uniform3-SAT class as shown in Figure 3. One difference is that local minima appear more prevalent for over-constrained problems from the HardSolvable3-SAT class than for the Uniform3-SAT class. The second difference is in the data for plateaus of level 1. The proportion of minima rises slightly from 380 clauses to 430 clauses, then dips sharply; hence there are more benches of level 1 in Figure 18 than in Figure 3. A detailed analysis of the reasons for the differences is beyond the scope of this paper, but the generation process seems to eliminate clause combinations that make for level 1 local minima at the expense of making more minima of higher levels. The higher percentages of local minima for over-constrained problem instances indicate that local search may have a harder time solving these problems.

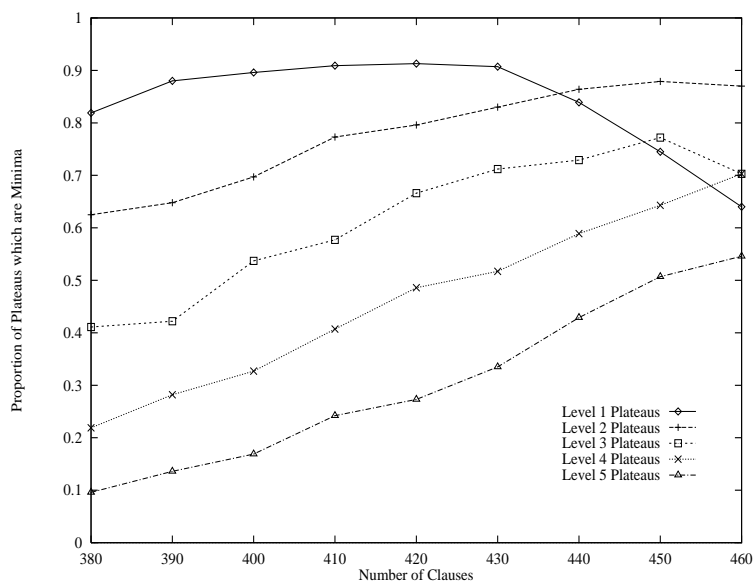

Figure 18: Proportion of Plateaus that are Local Minima vs. Number of Clauses for 100 Variable Problem Instances With Guaranteed Solutions

We analyzed the median size of local minima of different levels for the HardSolvable3-SAT class and found results similar to those reported for the Uniform3-SAT class in Figure 8. The median minima size for most levels of the HardSolvable3-SAT class are larger than those of the Uniform3-SAT class for under-constrained instances and somewhat smaller for over-constrained instances. We also found that the bench size distribution for the HardSolvable3-SAT class matched that of the Uniform3-SAT class. The median bench size is somewhat smaller for HardSolvable3-SAT instances than for Uniform3-SAT instances as the number of clauses in the problem instances grows. We also found that problems that were guaranteed to have solutions had a higher proportion of exits to bench size than randomly generated problems. It should not be surprising that this problem class is very similar to the previously investigated class since the problem generation algorithms are similar.

### 4.3 Cluster Problem Instances

We next present an analysis of plateaus for cluster problem instances; the generation procedure Cluster3-SAT is presented in Appendix B. These instances were created by generating 10 clusters of 10 variables and 34-40 clauses such that no variables are shared between clusters. The clusters are then linked with 20 connecting clauses such that each linking clause contains variables from distinct clusters. The total number of clauses in these instances ranges from 360 to 420. As in the earlier experiments, we guaranteed each instance had a solution using complete search. For each number of clauses in the clusters, we generated 1000 instances and again used GSAT to find plateaus at different levels. We found these instances took considerably longer to solve than the previous classes, similar to the results reported in Kask and Dechter (1995) (CPU times are not shown in this paper).

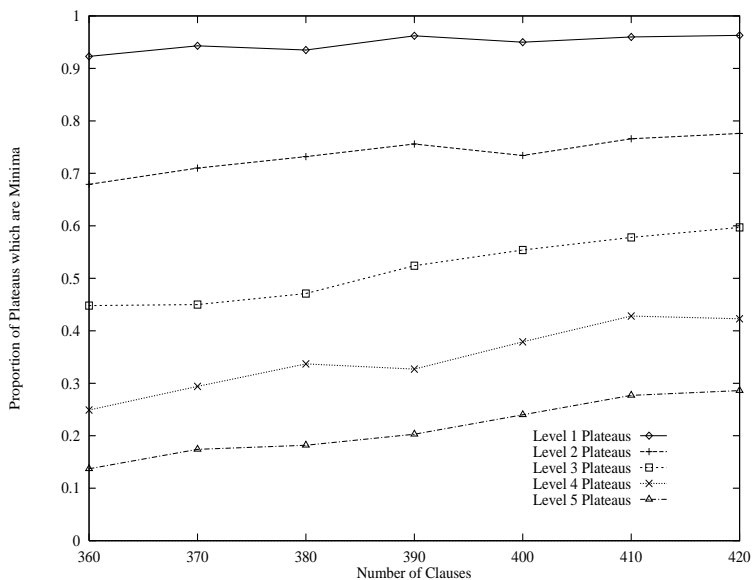

Figure 19: Proportion of Plateaus that are Local Minima vs. Number of Clauses for 100 Variable Cluster Problem Instances

Figure 19 shows the proportion of plateaus that are local minima graphed against the total number of clauses for cluster problem instances of 100 variables. The proportion of plateaus that are local minima are less dependent on the number of clauses in comparison to Figures 3 and 18. As a result there tend to be proportionally fewer local minima for over-constrained cluster problem instances in comparison to Uniform3-SAT instances. Figure 20 shows the median local minima size plotted against the number of clauses in the problem instances. The local minima for these problem instances are larger than the minima of Uniform3-SAT instances analyzed by a factor of about 5-10, as seen in Figure 8. We were unable to collect data for level 0 minima due to the excessive CPU requirements.

Figure 21 shows the median bench sizes plotted against the number of clauses in the problem instances. When compared to the median bench size of Uniform3-SAT instances in Figure 11, we see that the median size of benches for cluster problems is dramatically different. Cluster problem instances with fewer clauses per cluster resulted in enormous benches, in some cases larger than benches for Uniform3-SAT instances by a factor of 10.

The increase in the size of benches of cluster problem instances over randomly generated instances is accompanied by a decrease in the proportion of exits to bench size. Figure 22 shows the mean proportion of exits to bench size versus total number of clauses for cluster problem instances. In comparison to the same measure for Uniform3-SAT instances, shown in Figure 14, we see that benches for cluster problems have fewer exits than benches for Uniform3-SAT instances for all bench sizes from 1 to 5. The increase in bench size coupled with the decrease in exits means that local search is likely to have a much harder time escaping benches for cluster problems than for the other problem classes. This counters the good news that there are fewer minima for these problems.

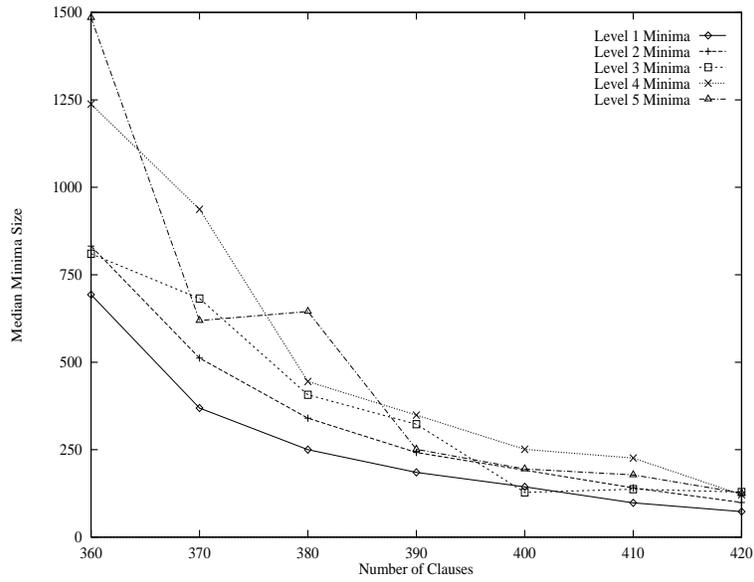

Figure 20: Median Size of Local Minima vs. Total Number of Clauses for 100 Variable Cluster Problem Instances

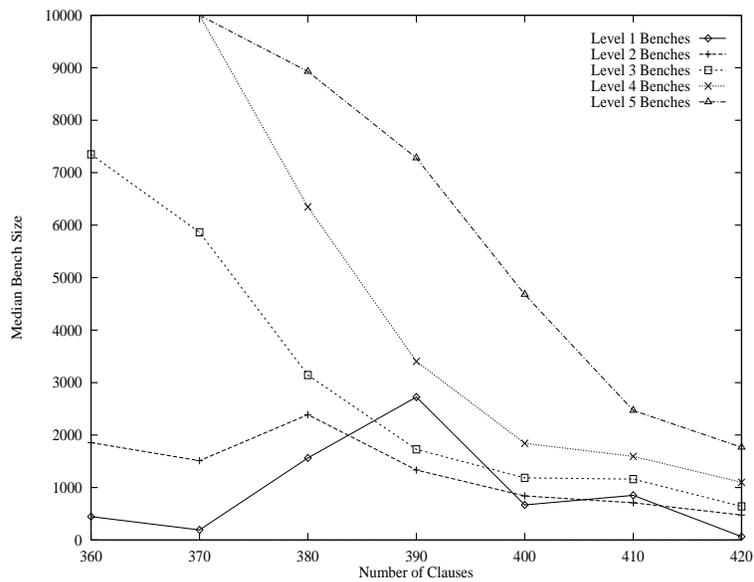

Figure 21: Size of Benches vs. Total Number of Clauses for 100 Variable Cluster Problem Instances

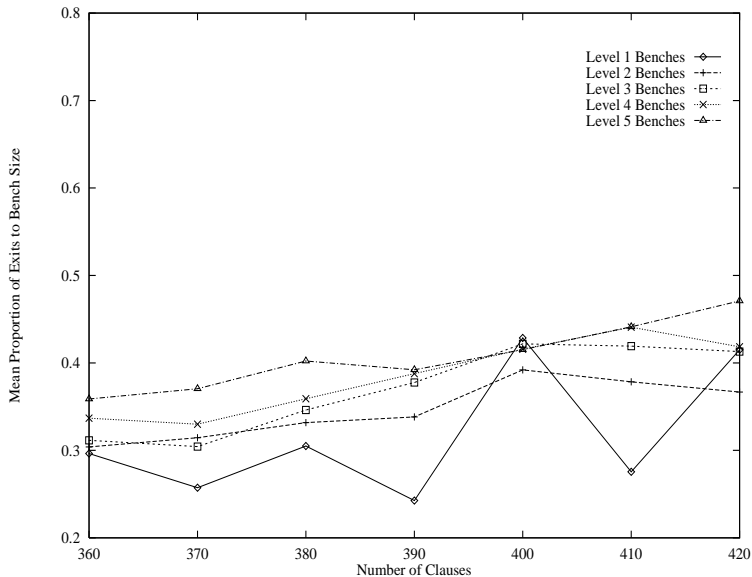

Figure 22: Mean Proportion of Exits to Bench Size vs. Size of Benches for 100 Variable Cluster Problem Instances

### 4.4 Summary

In summary, we see that the behavior of plateaus in unsatisfiable Uniform3-SAT problem instances differs from the behavior of satisfiable instances from the same class. Unsatisfiable instances have proportionally more local minima, smaller minima and smaller benches than satisfiable instances. Problems from the HardSolvable3-SAT class have more minima than those from the Uniform3-SAT class, and benches in the HardSolvable3-SAT class instances have more exits than benches in the Uniform3-SAT instances. As a result we expect problems in the HardSolvable3-SAT class to be harder for local search because local search algorithms will be more frequently trapped in local minima. Cluster problem instances have fewer local minima than Uniform3-SAT instances, but tend to have larger benches with fewer exits. We expect these problems to be harder because local search will have more trouble escaping benches.

### 5. Previous Work

Numerous researchers have studied local search techniques for $\mathcal{NP}$-Hard problems, addressing plateau behavior and local minima and how to escape them. However, the research has largely centered on analyzing the performance of the algorithms and less on the structure of the problem. Hence, improvements in algorithms are credited to a mechanism without a clear understanding of the properties of the problem that make the mechanism work. In this section we review some previous analysis of both algorithms and properties of local search spaces in light of our new discoveries. While this discussion focuses on GSAT and similar local search algorithms for the 3-SAT problem, we discuss the potential impact of our work on local search for other algorithms in the next section.

### 5.1 Analyzing Properties of Problem Spaces

Clark et al. (1996) studied how the number of solutions affected local search algorithm performance for both 3-SAT problems and for Constraint Satisfaction Problems. They showed that both the number of solutions and number of constraints play a role in determining how well GSAT works. Our work complements this study by adding to the understanding of how local search is affected by problem instance structure. We also add to their results by showing that solutions tend to occur in disconnected subgraphs of variable size.

Hampson and Kibler (1995) studied how plateaus affect local search's ability to solve 3-SAT problem instances and showed how a local search algorithm could be modified by performing complete search when a plateau is encountered. They analyzed the ratio of the number of exits off of benches versus the size of the benches for randomly generated problem instances at $\frac{C}{N}$=4.25. They found that benches at higher levels are easier to escape on the average, which is consistent with our findings. However, they failed to mention the high variance in the proportion of exits of benches, and also failed to report the existence of local minima. They also report that local search, augmented with complete search of plateaus, generally performed worse than the original local search. We believe that large plateaus, while rare, contributed to the large CPU times reported in their paper.

Gent and Walsh (1993a) investigated how GSAT solved problem instances by aggregating statistics of GSAT performance. They collected information on the number of satisfied clauses as a function of GSAT's flip number, the number of best flips as a function of flip number, and other statistics averaged over many runs and problem instances. Their study indicates that GSAT satisfies an average of 99% of the clauses after $2N$ flips on instances with $\frac{C}{N} = 4.3$. This works out to be 425 clauses for instances with 100 variables. Our evidence that there are local minima and hard-to-escape benches at level 5 (i.e., 425 satisfied clauses) is consistent with these results. Gent and Walsh also report that the number of flips GSAT spends on benches before escaping is highly variable during the second half of search, when the number of satisfied clauses is very high (Gent & Walsh, 1993a). This is consistent with our finding that benches can be either very easy or very hard to escape.

### 5.2 Revisiting Local Search Algorithms

When GSAT encounters a plateau it randomly searches the plateau. If the plateau is a bench, then GSAT can escape; however it may take a very long time if the bench has very few exits relative to its size, or if the exits are highly clustered in one region of the bench. If GSAT encounters a local minimum it will never escape; even if it made a move off the minimum to a state of higher level, GSAT will simply move back to the minimum, because every state on the minimum looks better than every state leading away from it. We should point out that GSAT can escape a local minimum of size 1 because it is forced to make a move. However, either GSAT will return immediately to this minimum or to another adjacent to the neighbor; as we found so few minima of size 1 and since GSAT doesn't exhibit such cycling behavior this is a minor consideration.

HSAT (Gent & Walsh, 1993b) augments GSAT with a heuristic designed to break ties. If HSAT has several flips that are equally good in terms of the number of satisfied clauses, then it flips the variable flipped longest ago. HSAT will explore benches more effectively than GSAT by ensuring that variables are flipped "fairly"; as long as HSAT remains on a bench it will not flip a variable that keeps HSAT on the bench until all other such variables are flipped. Therefore, HSAT's improved performance may be due to its ability to escape benches faster than GSAT. However, HSAT is still unable to escape local minima.

Tabu search (Glover, 1989; Mazure, Säis, & Grégoire, 1997) augments local search with a fixed length list of previously made moves. The algorithm is not allowed to reverse a move that is on the tabu list. Local search augmented with tabu lists may escape local minima. The memory structure will either explicitly or implicitly store states on the plateau and force local search to make moves away from that part of the space. However, due to the nature of the tabu list, it is possible that local search with one of these variants will ignore a move which reduces the objective function simply because it is on the Tabu list. This is because Tabu search frequently stores moves, not states. As a result, different tabu search implementations allow moves to states with fewer unsatisfied clauses than ever detected to date, even if the required move is on the tabu list, and thus can avoid this problem. This can result in tabu search missing exits from benches; whether this results in poor performance is unknown.

GSAT with random walk (Gent & Walsh, 1995) flips a randomly selected variable with probability $p$ and uses the standard criteria to select flips with probability $1-p$. This feature will allow GSAT to escape either local minima or benches, but does not guarantee that the next move will not simply bring GSAT back into the minima it escaped from This will not happen if there are multiple, equivalently good moves available. Gent and Walsh (1993a) report that, when the number of unsatisfied clauses is very small, the number of available moves leading to a reduction in level for GSAT tends to be 1. However, the effectiveness of random walk suggests that a random flip will move GSAT into a place where it can proceed to a solution. Also, if $p$ is large, then two random walk steps can follow each other in succession, improving the chances of escaping local minima. The fact that this variant results in substantial improvements even when used with other modifications to GSAT such as tie-breaking heuristics (Gent & Walsh, 1995) complements our discovery in Section 3 that local minima tend to be shallow; random walk may effectively promote escape from these local minima into other parts of the search space.

WalkSAT (Kautz & Selman, 1996) does not examine all possible neighbors before moving. Instead, WalkSAT randomly selects an unsatisfied clause and only investigates states reachable by flipping variables in the selected clause. As a result the neighborhood examined changes from flip to flip, and the reverse move may not be in the next neighborhood examined. WalkSAT performs much of its search blind to the features we have uncovered. WalkSAT can escape local minima by simply not choosing a neighborhood containing moves back onto the minima, or it may take a series of worse moves to escape a bench with many exits simply because its neighborhood did not contain them.

Simulated annealing (Kirkpatrick, Gelatt, & Vecchi, 1983) only examines a single neighbor of the current assignment. Moves leading to improvements in the objective function are always accepted, while moves that worsen the objective function are accepted probabilistically; the probability is based on how much worse the move is and how long the search has progressed. Like WalkSAT, simulated annealing conducts much of its search blind to plateau features. Simulated annealing can make a backwards move on a bench or minimum even if a neighboring state results in a forward move; while this can help escape minima and large benches, it may be a sub-optimal strategy early in search.

## 6. Next Generation Local Search Algorithms

We have identified and analyzed a number of features of local search topology that may influence the success of local search. How can our results be used to improve local search algorithms? One contribution of this study is to identify features of the local search space that are worth investigating before beginning construction of a local search algorithm to solve a new problem. A rapid exploration of the properties of benches and local minima can be undertaken to determine which local search tactics are likely to work best for this search problem class. For instance, such an examination might reveal that for one problem local minima are very prevalent but uniformly small, indicating that explicit local minima detection and avoidance is likely to be an effective tactic. Also, it is possible to use results such as those in Figure 3 to determine an adaptive schedule for resetting the probability of random walk or optimizing the size of the tabu list, as is done in Mazure et. al. (1997). It is also possible that new classes of local search algorithms could *learn* which tactics work best in a manner similar to MultiTac (Minton, 1996). Our study provides a first step towards identifying features that should be tracked by these self-modifying local search algorithms.

When a local search algorithm starts exhibiting plateau behavior, it may be on a small minima, a large minima, a bench with many exits, or a bench with few exits. (We ignore the case of a small bench, since it is not too hard to escape in these cases.) The problem is to identify which case the search process is stuck in, and then choose the proper tactic to handle it. Standard tactics include continuing search as normal, invoking a special detection procedure, randomly flipping one variable as in random walk, randomly flipping a small number of variables as in Jump-SAT (Gent & Walsh, 1995) or randomly generating new values for all variables as in randomly restarting.

Small minima can be detected easily using an algorithm such as Breadth-First Search, as was done by Hampson and Kibler (1995). Once a local search algorithm has detected and escaped a local minima, it is desirable to prevent it from revisiting the minima it has escaped. Local search could proceed by "filling in" local minima as they are found in order to prevent revisiting them. This is approximately how tabu search works, and other schemes can be used as well. The small size of most local minima indicates that memory requirements for such a scheme are small as long as only small numbers of minima are encountered. An algorithm using this mechanism could then explore numerous local minima that are close together in the solution space without restarting.

Large benches and minima are more difficult to recognize and escape. The question becomes one of determining the utility of continuing to search versus changing tactics. The studies we have done provide algorithm designers the information required to implement the utility computation so that a local search algorithm can intelligently choose from among its tactics. For instance, knowing that a problem instance is a cluster problem is indicative that large benches with few exits are more likely to inhibit local search than local minima. Hence explicit local minima detection is not a good strategy for this problem class; jumping or random restart might be a better strategy.

We should point out that while many enhancements like those proposed above are in place for local search algorithms to solve 3-SAT problems, these enhancements have not been applied to other problem classes such as Graph k-Coloring. We expect these extensions to be successful at improving the performance of local search algorithms to solve these problems. One way to approach new problems is to spend time gathering information on the topology of the search space, as we have done in this paper. A second option, as we mentioned above, is to use knowledge of the local search topology to learn the best tactics while solving instances. Detailed information on the appearance of local minima, distribution of local minima size, bench size, and prevalence of exits from benches can then be used to construct very good local search procedures that explicitly take these factors into account.

## 7. Conclusions and Future Work

We have presented an analysis of important properties of plateau structures in local search spaces that can be used by local search algorithm designers to construct better local search algorithms. We have defined a set of topological structures of local search spaces and shown how they affect local search. We have provided conclusive evidence of the existence of local minima in search spaces, and shown that they become more prevalent as the number of unsatisfied clauses becomes close to 0. We have also shown that plateau behavior in local search is caused by both local minima and benches. Our results show that both local minima and benches vary widely in size; while both are most often small, large local minima and benches may defy detection and avoidance by local search algorithms. We also show how the characteristics of these structures change with different problem classes. Our analysis has made it possible to interpret previous work on improving local search in terms of the search space structure, illuminating the importance of escaping benches early in search and detecting local minima later in search.

We have made suggestions in the previous section that might be used to create new versions of local search that are better than the current crop of algorithms. An obvious next step is to implement these algorithms and analyze their performance, especially in comparison to existing algorithms that already attempt to escape plateaus.

We have barely begun analyzing the topology of plateaus. While we have some empirical evidence that local minima of low level (i.e., near 0) can be escaped by unsatisfying only one additional clause, this may not be true for more structured problem classes. The evidence that benches may have many exits does not always imply they are easy to escape; highly clustered exits may make benches hard to escape. Further analysis of the topology of plateaus for a variety of problem instances will lead to more concrete results that can influence local search algorithm development.

It is clear that the nature of plateaus is highly dependent on the problem class being tested. Extending this form of analysis to other problem classes might reveal differences in plateau structure that motivate substantially different GSAT variants. Furthermore, plateaus in Graph Coloring Problem and the Traveling Salesperson Problem may manifest themselves in different ways than those for 3-SAT, and local search algorithms for these problems will explore plateaus differently. These differences must be studied in order to determine how best to apply a new understanding of local search topology. It is also unclear how this study will extend to such problems as Traveling Salesperson Problem where the search space may be much "smoother."

While we have collected a large amount of data on local search spaces, we have not had much success in modeling the features we defined. While it is possible to compute the probability that an individual state in a search space is on a local minima or is on a bench with exits, it is very difficult to compute the expected size of a bench or minimum without making horrendous independence assumptions. Further work on such modeling may result in a better understanding of local search topology.

As mentioned in Section 2, it is possible to alter our definition of benches to specifically exclude contours as a type of bench. The rationale is that contours provide no impediment to greedy local search, and very little impediment to the semi-greedy variations. One possibility is to change the definition of plateau to only include states without neighbors of both higher and lower level. There are several predictable effects of this change. First, we know that some of our reported benches are pure contour regions. These would be eliminated from consideration in reporting proportions of plateaus that are benches. Second, the size of the benches would exclude these states, and so we expect to find smaller benches under the exclusive bench definition. Third, the measurements of the mean proportion of states on benches that are exits would also change, because excluding contours reduces the mean. Fourth, these states provide a potential means of linking bench regions that would be disjoint under the exclusive definition. Thus it is possible that use of the exclusive definition will cause a dramatic reduction in average bench size, accompanied by an increase in bench numbers. It might even eliminate the large size tails in the bench size distributions that we observed using the inclusive definition. This would significantly alter our conclusions regarding how benches impede local greedy search, and our recommendations regarding how to deal with such benches. Exploring the impact of such revised definitions in the explanation of plateau behavior is worth investigating.

Finally, our analysis of topological structures is geared towards analyzing local search algorithms with greedy objective functions based on the number of unsatisfied constraints. Many local search algorithm designers are experimenting with new objective functions that are modified continuously throughout search, as in clause weighting schemes (Selman & Kautz, 1993). Work of this type may lead to more innovations in the design of objective functions. Since plateau behavior is rooted in the objective function used, our analysis is not suitable for analyzing these methods, but may provide insight into how to conduct a similar study for self-modifying algorithms of this type.

## Acknowledgements

We gratefully acknowledge the comments of the JAIR reviewers and editors; we also acknowledge the comments of Phil Rogaway and Chip Martel, both of U.C. Davis.

## Appendix A. Sample Problem

This section illustrates some of the terms defined in Section 2. Consider the following 4 variable, 14 clause 3-SAT problem instance:

$$(A \vee B \vee C) \wedge \quad (A \vee B \vee \overline{C}) \wedge \quad (A \vee \overline{B} \vee C) \wedge$$
$$(A \vee \overline{B} \vee \overline{C}) \wedge \quad (\overline{A} \vee B \vee C) \wedge \quad (\overline{A} \vee B \vee \overline{C}) \wedge$$
$$(\overline{A} \vee \overline{B} \vee C) \wedge \quad (\overline{A} \vee \overline{B} \vee D) \wedge \quad (A \vee B \vee \overline{D}) \wedge$$
$$(A \vee \overline{C} \vee D) \wedge \quad (A \vee \overline{B} \vee D) \wedge \quad (\overline{A} \vee B \vee D) \wedge$$
$$(A \vee C \vee \overline{D}) \wedge \quad (\overline{A} \vee C \vee D)$$

For the duration of this section we will abbreviate assignments of values to variables in the following way: 0 is False, 1 is True, and hence a string of 0s and 1s of length 4 encodes an assignment to the variables ABCD in order.

This problem instance has a global minimum comprised of a single solution at 1111. The single state 0000 is local minimum of size 1 and level 1, i.e., has one unsatisfied clause. The border of this local minimum consists of the states 1000,0100,0010,0001; states 0001 and 1000 have level 3 and the other two states have level 2.

The following states constitute a bench of level 1: 1001, 1101 and 1011. 1101 is an exit since flipping C results in 1111, the solution. Similarly, 1011 is also an exit since flipping B results in the solution. The neighbors of 1001 that are not on the bench are 0001 and 1000; each of these has level three, so 1001 is not an exit.

| State | Comment | Level | Unsatisfied Clauses |
|---|---|---|---|
| 1111 | Solution | 0 | |
| 0000 | Local Minimum | 1 | $(A \vee B \vee C)$ |
| 0010 | Border of Minimum | 2 | $(A \vee B \vee \overline{C}), (A \vee \overline{C} \vee D)$ |
| 0100 | Border of Minimum | 2 | $(A \vee \overline{B} \vee C), (A \vee \overline{B} \vee D)$ |
| 0001 | Border of Minimum | 3 | $(A \vee B \vee C), (A \vee B \vee \overline{D}), (A \vee C \vee \overline{D})$ |
| 1000 | Border of Minimum | 2 | $(\overline{A} \vee B \vee C), (\overline{A} \vee B \vee D)$ |
| 1001 | Bench | 1 | $(\overline{A} \vee B \vee C)$ |
| 1011 | Bench | 1 | $(\overline{A} \vee B \vee \overline{C})$ |
| 1101 | Bench | 1 | $(\overline{A} \vee \overline{B} \vee C)$ |
| 1111 | Border of Bench | 0 | |
| 0001 | Border of Bench | 3 | $(A \vee B \vee C), (A \vee B \vee \overline{D}), (A \vee C \vee \overline{D})$ |
| 1000 | Border of Bench | 2 | $(\overline{A} \vee B \vee C), (\overline{A} \vee B \vee D)$ |
| 0010 | Contour | 2 | $(A \vee B \vee \overline{C}), (A \vee \overline{C} \vee D)$ |
| 1010 | Contour | 2 | $(\overline{A} \vee B \vee D), (\overline{A} \vee B \vee \overline{C})$ |
| 0011 | Contour | 2 | $(A \vee B \vee \overline{C}), (A \vee B \vee \overline{D})$ |
| 0000 | Border of Contour | 1 | $(A \vee B \vee C)$ |
| 1110 | Border of Contour | 1 | $(\overline{A} \vee \overline{B} \vee D)$ |
| 0111 | Border of Contour | 1 | $(\overline{A} \vee \overline{B} \vee C)$ |

Figure 23: Some Topological Structures of the Sample Problem Instance

The states 0010, 1010 and 0011 form a level 2 bench which also is a contour. 0010 is a neighbor of the local minimum at level 1, 1010 is adjacent to 1110 which is at level 1, and 0011 is adjacent to 0111 which is at level 1.

The states 1000 and 1100 form a bench of level 3 which is a contour. Each of the states 0110, 0001 are also contours of level 3 by themselves. Since there are no states unsatisfying more than three clauses these contours are in fact local maxima.

These features are summarized in Figure 23.

## Appendix B. Random Problem Generation

This appendix contains pseudo-code for the random problem classes studied in this paper. First we present the Uniform3-SAT class. Parameters to this generator are $C$ the number of clauses and $N$ the number of variables. In this class the procedure selects three literals without replacement from $N$ and assigns each a negative sign with probability $\frac{1}{2}$. This procedure was first presented in Crawford and Auton (1993) and appears in Figure 24.

```
procedure Uniform3-SAT(C,N)
    Σ = ∅
    for (i=1 to C)
        Clause= 3 distinct variables selected uniformly from 1..N
        Negate each variable in Clause with probability ½
    Σ = Σ ∪ Clause
    end for
    return Σ
end
```

Figure 24: Random Problem Generation Algorithm Sketch

Next we present the Cluster3-SAT problem generator. The parameters are the number of clauses $C$, the number of variables $N$ per cluster, the number of clusters $M$, and the number of linking clauses $L$. This generator builds instances by first creating $M$ independent sub-problems of $C$ clauses and $N$ variables each, using the Uniform3-SAT generation procedure described above. The variables of these sub-problems are relabeled so that no sub-problem shares variables with any other sub-problem; sub-problems are then linked by generating $L$ linking clauses. Each linking clause contains variables from three distinct sub-problems. Kask and Dechter generate these problems using the HardSolvable3-SAT procedure defined below as described in Kask and Dechter (1995). The procedure appears in Figure 25.

Finally we present the HardSolvable3-SAT generator. The parameters are the number of clauses $C$ and the number of variables $N$. Instances are generated by first selecting an assignment $S$ to be a guaranteed solution. Clauses are generated as in Uniform3-SAT, however if a clause has either zero or two satisfied literals under the selected assignment it is rejected. For instance, the clause $(A \lor B \lor C)$ would be rejected if the assignment $S$ was 110 since it has two satisfied literals under this assignment. This preserves a uniform balance of signs for each variable in the limit, resulting in little information about the solution being present in sign balances in the problem instance. This method is discussed further in Tsuji and Van Gelder (1993) and the algorithm is given in Figure 26.

```
procedure Cluster3-SAT(C,N,M,L)
    # First generate M sub-problems with distinct variables
    for i=1 to M
      Γ_i=Uniform3-SAT(Γ,C,N)
      Re-label literals in Γ_i so that no sub-problem shares variables
      Σ = ∪_{i=1}^{M} Γ_i
    end for
    # Next generate linking clauses
    for i = 1 to L
      Randomly select 3 distinct sub-problems Γ_a, Γ_b, Γ_c from the Γ_i s
      Clause= one variable randomly selected from each of Γ_a, Γ_b, Γ_c
      Negate each variable in Clause with probability 1/2
      Σ = Σ ∪ Clause
    end for
    return Σ
end
```

Figure 25: Cluster Problem Generation Algorithm Sketch

```
procedure HardSolvable3-SAT(C,N)
    Σ = ∅
    S = randomly generated assignment to the variables
    while (i < C)
      Clause= 3 distinct variables selected uniformly from 1..N
      Negate each variable in Clause with probability 1/2
      # Check to make sure Clause allowed in formula under S
      if (1 or 3 literals of Clause true under S)
        Σ = Σ ∪ Clause
        i++
      end if
    end while
    return Σ
end
```

Figure 26: "Hard" Guaranteed Satisfiable Problem Generation Algorithm Sketch

A final note on random problem instance generation is in order. None of these procedures guarantees that the resulting problem instance will contain all the variables. If a large number of variables and a small number of clauses are used as parameters, then the resulting problem may not contain all variables. However, for the ranges of clauses and variables used in this work all problem instances had the full range of variables.

# References

Cheeseman, P., Kanefsky, B., & Taylor, W. (1991). Where the *really* hard problems are. *12th International Joint Conference on Artificial Intelligence*, 163–169.

Clark, D., Frank, J., Gent, I., MacIntyre, E., Tomov, N., & Walsh, T. (1996). Local search and the number of solutions. *Proceedings of the 2d International Conference on Principles and Practices of Constraint Programming*, 119–133.

Crawford, J., & Auton, L. (1993). Experimental results on the crossover point in satisfiability problems. *Proceedings of the 11th National Conference on Artificial Intelligence*, 21–27.

Gent, I., & Walsh, T. (1993a). An empirical analysis of search in GSAT. *Journal of Artificial Intelligence Research*, *1*, 47–59.

Gent, I., & Walsh, T. (1993b). Towards an understanding of hill-climbing procedures for SAT. *Proceedings of the 11th National Conference on Artificial Intelligence*, 28–33.

Gent, I., & Walsh, T. (1995). Unsatisfied variables in local search. In Hallam, J. (Ed.), *Hybrid Problems, Hybrid Solutions*, pp. 73–85. IOS Press.

Glover, F. (1989). Tabu search part I. *ORSA Journal on Computing*, *1*(3), 190–206.

Hampson, D., & Kibler, S. (1995). Large plateaus and plateau search in boolean satisfiability problems: When to give up searching and start again. In Johnson, D., & Trick, M. (Eds.), *DIMACS Series in Discrete Mathematics and Theoretical Computer Science: Cliques, Colors and Satisfiability*, Vol. 26, pp. 437–456. American Mathematical Society.

Kask, K., & Dechter, R. (1995). GSAT and local consistency. *Proceedings of the 14th International Conference on Artificial Intelligence*, 616–622.

Kautz, H., & Selman, B. (1996). Pushing the envelope: Planning, propositional logic and stochastic search. *Proceedings of the 13th National Conference on Artificial Intelligence*, 1194–1201.

Kirkpatrick, S., Gelatt, C., & Vecchi, M. (1983). Optimization by simulated annealing. *Science*, *220*(4598), 671–680.

Mazure, B., Säis, L., & Grégoire, E. (1997). Tabu search for GSAT. *Proceedings of the 14th National Conference on Artificial Intelligence*, 281–286.

Minton, S. (1996). Automatically configuring constraint satisfaction programs: A case study. *Constraints*, *1*(2), 7–43.

Selman, B., & Kautz, H. (1993). Domain independent versions of GSAT: Solving large structured satisfiability problems. *13th International Joint Conference on Artificial Intelligence*, 290–295.

Selman, B., Levesque, H., & Mitchell, D. (1992). A new method for solving hard satisfiability problems. *Proceedings of the 11th National Conference on Artificial Intelligence*, 440–446.

Tsuji, Y., & Gelder, A. V. (1993). Incomplete thoughts about incomplete satisfiability procedures. *Proceedings of the 2d DIMACS Challenge*.